# Artificial Intelligence and Systems Theory: Applied to Cooperative Robots

Pedro U. Lima & Luis M. M. Custodio
Institute for Systems and Robotics
Instituto Superior Técnico
Av. Rovisco Pais, 1
1049-001 Lisboa, PORTUGAL
{pal,lmmc}@isr.ist.utl.pt

*Abstract:* This paper describes an approach to the design of a population of cooperative robots based on concepts borrowed from Systems Theory and Artificial Intelligence. The research has been developed under the SocRob project, carried out by the Intelligent Systems Laboratory at the Institute for Systems and Robotics - Instituto Superior Técnico (ISR/IST) in Lisbon. The acronym of the project stands both for "Society of Robots" and "Soccer Robots", the case study where we are testing our population of robots. Designing soccer robots is a very challenging problem, where the robots must act not only to shoot a ball towards the goal, but also to detect and avoid static (walls, stopped robots) and dynamic (moving robots) obstacles. Furthermore, they must cooperate to defeat an opposing team. Our past and current research in soccer robotics includes cooperative sensor fusion for world modeling, object recognition and tracking, robot navigation, multi-robot distributed task planning and coordination, including cooperative reinforcement learning in cooperative and adversarial environments, and behavior-based architectures for real time task execution of cooperating robot teams.
*Keywords*: multi-robot systems, sensor fusion, distributed task planning, robot navigation, soccer robots.

## 1. Introduction

Cooperative Robotics is a modern research field, with applications to areas such as building surveillance, transportation of large objects, air and underwater pollution monitoring, forest fire detection, transportation systems, or search and rescue after large-scale disasters. In short, a population of cooperative robots behaves like a distributed robot to accomplish tasks that would be difficult, if not impossible, for a single robot. Many lessons important for this domain can be learned from the Multi-Agent Systems field of Artificial Intelligence (AI) concerning relevant topics for Cooperative Robotics, such as distributed continual planning (desJardins, M. E., *et al*, 1999), task allocation (Ferber, J., 1999), communication languages or coordination mechanisms (Decker, K. S., & Lesser, V. R., 1995). Robotic soccer is a very challenging problem, where the robots must cooperate not only to push and/or kick an object (a ball) towards a target region (the goal), but also to detect and avoid static (walls, stopped robots) and dynamic (moving robots) obstacles while moving towards, moving with or following the ball. Furthermore, they must cooperate to defeat an opposing team. All these are features common to many other cooperative robotics problems. This paper surveys the several research problems addressed by the SocRob project, building a Systems Theory standpoint on AI concepts. In Section 2 we describe our view of the general problem involving multiple robots that act as a team, cooperating and coordinating their actions to attain the team goal. Needless to say, single-robot "traditional" research problems are covered, both from the sub-system and the integration standpoints. Natural extensions to cooperative multi-robot teams are also detailed. The problems addressed so far and the solutions we obtained for them are described in Section 3. Open problems of interest for the project and clues on how we intend to approach their solution are discussed in Section 4. We end the paper drawing some conclusions in Section 5.

## 2. A General Multi-Robot Cooperation and Coordination Problem

Many researchers around the world are designing mobile robots capable to display increasing autonomy and machine intelligence properties. Most groups concentrate in specific subsystems of a robot, such as the planner, the navigator, or the sensor fusion. What usually is missing in their design is a systematic way to glue together all these subsystems in a consistent fashion. Such a methodology, should one be available, would help engineering the mobile robots of the future.

One of the key factors of success for a robot lies on its capability to perceive correctly its surrounding environment, and to build models of the environment adequate for the task the robot is in charge of, from the information provided by its sensors. Different sensors (e.g., vision, laser, sonar, encoders) can provide alternative or complementary information about the same object, or information about different objects. *Sensor fusion* is the usual designation for methods of different types to merge the data from the several sensors available and provide improved information about the environment (e.g., about the geometry, color, shape and relevance of its objects). When a team composed of several cooperating robots is concerned, the sensors are spread over the different robots, with the important advantage that the robots can move (thus moving its sensors) to actively improve the cooperative perception of the environment by the team. The information about the environment so obtained can be made available and regularly updated by different means (e.g., memory sharing, message passing, using for instance wireless communications) to all the team robots, so as to be used by the other sub-systems.

Once the information about the world is available, one may think of using it to make the team behave autonomously and machine-wise intelligently. Three main questions arise for the team:

Where and which *a priori* knowledge about the environment, team, tasks and goals, and perceptual information gathered from sensors, should be kept, updated and maintained? This involves the issue of distributed knowledge representation adequate to consistently handle different and even opposite views of the world.

What must be done to achieve a given goal, given the constraints on time, available resources and distinct skills of the team robots? The answer to this should provide a team *plan*.

How is the actual implementation of a plan handled, ensuring the consistency of individual and team (sub)-goals and the coordinated execution of the plan?

So far, a *bottom-up* approach to the *implementation* of a cooperative multi-robot team has been followed in the SocRob project, starting from the development of single robot sub-systems (e.g., perception, navigation, decision-making) and moving towards relational behaviors, comprehending more than one robot.

However, a key point is a *top-down* approach to system *design*. The design phase establishes the specifications for the system: qualitative specifications - concerning formal logical task design so as to avoid deadlocks, livelocks, unbounded resource usage and/or sharing non-sharable resources, and to choose the primitive tasks that will span the desired task space;

quantitative properties - concerning performance features, such as accuracy (e.g., the spatial and temporal resolution, as well as the tolerance interval around the goal, at each abstraction level), reliability and/or minimization of task execution time given a maximum allowed cost.

To support this *top-down design* and *bottom-up implementation* philosophy, suitable functional and software architectures, respectively, must be conceived prior to the development of all the sub-systems.

*2.1. Single-Robot Research Problems*

Most of the problems tackled so far within the SocRob project concern the sub-systems of the individual robots composing a team. From our standpoint, relevant topics are:

**Functional and Software Architectures**: Modern robots should be designed based on a top-down design from specifications to ensure desired performance levels (both qualitative and quantitative). Therefore, the designers should start by specifying a functional architecture which will guide the design of the robot sub-systems in an integrated fashion, i.e., each sub-system is not necessarily designed to optimize its performance but rather aiming at optimizing the overall system performance. Another important issue is to determine, given the desired task space (i.e., the set of tasks that will have to be carried out by the robot in a particular application), the minimal set of primitive tasks that will span that task space. Moreover, the final implementation should be supported on a suitable software architecture designed to allow real-time multi-processing, data sharing and mutually exclusive allocation of shared resources among the robot sub-systems.

**Single-Robot Task Planning**: Given the primitive task set referred in the previous item, the robot must be able, given the current and past world states (including its own internal state), to compose primitive tasks so as to come up with a plan that carries out a given desired task. There may be more than one plan that accomplishes a task, but *a posterior* decision system should be able to determine, eventually based on machine learning, the one that achieves the best performance, based on the available information and prediction horizon.

**Single-Robot Task Coordination**: Plans must be such that they allow continuous handling of the environment uncertainties and unexpected events. Once a plan is determined, task coordination deals with its execution. Plan execution must, at least, take into account the detection of events, smooth transitions between primitive tasks, synchronization of primitive tasks executed concurrently, mutual exclusion when two or more tasks attempt to access shared resources, iterative estimation of primitive task performance, learning how to improve a plan over time by choosing more convenient algorithms among those available for each primitive task, and so on.

**Navigation**: The navigation system is an important sub-system of a mobile robot. In many applications one important feature of the navigation system concerns the ability of the robot to self-localize, i.e., to autonomously determine its position and orientation (posture). Using posture estimates, the robot can move towards a desired posture, i.e., by following a pre-planned virtual path or by stabilizing its posture smoothly (Canudas de Wit, C., *et al*, 1996). If the robot is part of a cooperative multi-robot team, it can also exchange the posture information with its teammates so that appropriate relational and organizational behaviors may be established. In robotic soccer, these are crucial issues. If a robot knows its posture, it can move towards a desired posture (e.g., facing the goal with the ball in between). It can also know its teammate postures and prepare a pass, or evaluate the game state from the team locations. Most approaches to Navigation determine with high accuracy the posture of the robot with respect to a given coordinate frame. However, this approach is typically resource-consuming, requiring the robot to spend a significant percentage of its processing time with the navigation sub-system, disregarding other important sub-systems, such as perception or planning, to name but a few. Furthermore, high accuracy is not always required for navigation purposes. One may be just interested to move closer to an object, rotate to see a given landmark, or move to another region. In those cases, another approach to navigation, known as topological (or relative) navigation, is advisable.

**Object Recognition and Tracking Using Sensor Fusion**: The ability to discriminate and recognize its surrounding objects, to distinguish the relevant ones and to track, among them, those that are relevant, is a major problem for any robot. For soccer robots, this problem is simplified since the relevant objects are distinguished by their colors (e.g., the ball is orange, the goals are blue and yellow). Nevertheless, fast and reliable color segmentation is not a trivial problem and requires some attention too. Furthermore, object detection may be performed by more than one sensor, such as different virtual sensors based on the vision transducer (e.g., mass center, edge detector, color segmentation), sonars, infrared and others. Therefore, sensor fusion arises as an important topic.

*2.2. Cooperative Multi-Robot Research Problems*

**Functional and Software Architectures**: If a team of cooperative robots is involved, the single-robot architectures of each of the team members must be integrated in the overall team architecture. The most usual solutions concerning the software architecture are: *centralized*, where one of the robots (or an external machine) processes the data acquired and sent by all the team members, takes all the team decisions and sends commands to the others; *distributed*, where local data processing is made at each of the robots but then information is sent to one of them to take the decisions; *fully decentralized*, where each robot takes its own decisions based on its own data and on information exchanged with its teammates.

The functional architecture of a behaviour-based multi-robot team must also classify behaviours according to their functionality. One such division consists of considering *organizational*, *relational* and *individual* behaviours (Drogoul, A., and Collinot, A., 1998), further described below.

**Multi-Robot Task Planning and Allocation**: In the multiple-robot case, plans must take into account the distributed nature of the task at hand. Different tasks must be allocated to the different robots in the team, according to their skills and performance. So, the planning and allocation system must be able to establish (sub)groups of robots within a team, and the robots must have and know how to deal with the notion of "belonging to a group". Therefore, plans must also include synchronization and communication among team members involved in the task. Moreover, if a robot cannot fulfill its assigned task, the task may simply be re-assigned to a robot within the group, a new robot may be integrated in the group to perform that task, or in the worst case a re-planning strategy has to be applied.

**Multi-Robot Task Coordination**: The extension of task coordination to a team of multiple robots introduces issues related to knowledge distribution and maintenance, as well as communications and related problems (e.g., noise, protocols, limited bandwidth). Furthermore, communication can be explicit (e.g., through wireless radio-frequency channels) or implicit (e.g., through the observation of teammate actions, should an *a priori* model of the teammates behaviour exist). The coordination of a task carried out by a team of cooperating robots involves signalling events detected by one robot which are relevant for some or all of its teammates and/or to exchange information obtained locally by the different robots of the team. Whenever a formation is required, several formation topologies are possible and the one suitable for the task at hand must be chosen as part of the coordination process. Although not inevitable, communications among team members are also required to keep the formation under control.

When the population is composed of heterogeneous robots, if a robot has to perform a particular task for which it does not have the necessary skills, it may ask another robot with the adequate skills to carry it out. In the particular case of the SocRob robotic team, where the robots are homogeneous, examples of cooperative behaviour are the cooperative localization of the ball, the execution of a pass, the dynamical exchange of player roles or the decision of which robot should go for the ball. All of them require some form of inter-robot coordination and underlying teamwork methodologies.

**Distributed World Modeling**: A team composed of multiple robots, possibly heterogeneous concerning on board sensing, can benefit from the availability of a world model, obtained from the observations made by the different team members and its on board sensors. This world model can be richer that if it were obtained by a single robot, due to the coverage of a broader area by a more diversified set of sensors. It can also be distributed through the teammates, e.g., by keeping in a single robot information which is only relevant locally and by broadcasting information gathered locally but which is of interest for the team as a whole. The sensor fusion problem is similar to the single-robot case, with the important difference that the sensor subsets are now independently mobile and can be actively positioned to improve the determination of object characteristics.

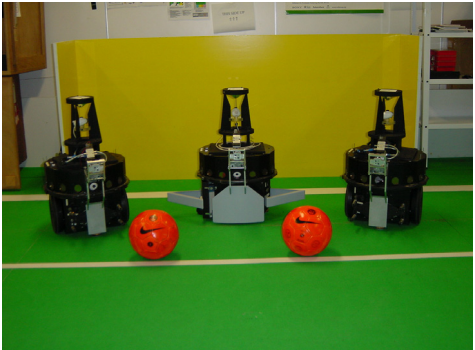

Fig.1 – Three robots of the current SocRob team.

## 3. Problems Already Addressed

A key issue of the research work developed under the SocRob project is the application of conceptual results to real robots participating in the Middle Size League (MSL) of RoboCup. The current robot team, displayed in Fig. 1, is composed of 4 Nomadic Super Scout II commercial platforms, later significantly modified by our group, each of them including: two-wheel differential drive kinematics, sixteen sonar sensors radially distributed around the robot, equally spaced, Motorola MC68332 based daughter board with three-axis motor controller, sonar and bumper interface, and battery level meters, two 12V batteries, 18Ah capacity, Pentium III 1000MHz based motherboard, with 512MB RAM, 8GB disk, two Philips USB WebCam 740K Pro, IEEE 802.11b wireless Ethernet PCMCIA card, pneumatic kicking device, based on Festo components, plus one bottle for pressurized air storage. In the remaining subsections, we describe some of the research problems addressed and solved for this team of robots.

*3.1. Color Segmentation and Cooperative Object Recognition*

A color segmentation interface was developed, providing two alternatives to discriminate the relevant MSL colors in HSV (Hue-Saturation-Value) color space (Gonzalez, R., & Woods, R., 1992): adjusting HSV intervals and graphically selecting regions with a given pixel color. The two approaches are cumulative. Furthermore, object segmentation is a topic directly related to the previous one, as we discriminate objects, namely the ball and the goals, not only based on their color, but also on their shape (e.g., by fitting circles to observed orange bulbs and identifying the ball with the closest and more circular bulb). A topic of current research within the project is the use of sensor fusion for world modeling. The goal is to maintain and update over time information on the relevant objects, such as ball position and velocity, teammates pose and velocity, opponents pose and velocity, or position of the goals with respect to the robot. Such information is obtained by each robot from the observations of its front and up cameras and then fused among all the team robots (Pinheiro, P. & Lima, P., 2004), using a Bayesian approach to sensor fusion, as depicted in Fig. 2. Currently this approach is used to provide information on ball position to all the team members, therefore enabling robots that do not see the ball to know where it is, besides improving ball localization reliability. Fusion is not used when two robots disagree (in probabilistic terms) on the ball localization.

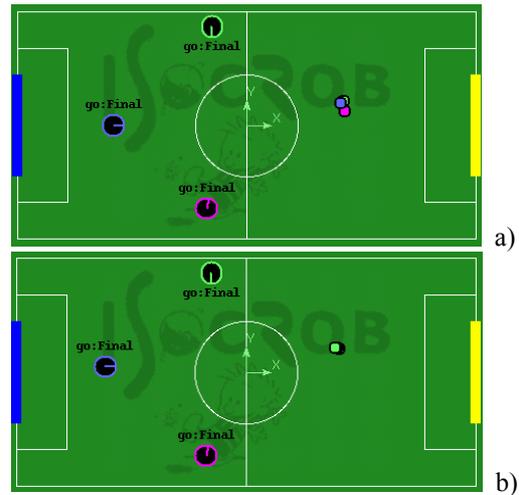

Fig. 2 – a) local (internal to each robot) sensor fusion enabled and global (among team robots) sensor fusion disabled; b) both local and global sensor fusion enabled

*3.2. Vision-Based Self-Localization*

An algorithm that determines the posture of a robot, with respect to a given coordinate system, from the observation of natural landmarks of the soccer field, such as the field lines and goals, as well as from *a priori* knowledge of the field geometry, has been developed within the SocRob project (Marques, C., & Lima, P., 2001). The algorithm is a particular implementation of a general method applicable to other well-structured environments, also introduced in (Marques, C., & Lima, P., 2001). The landmarks are processed from an image taken by an omni-directional vision system, based on a camera plus a convex mirror (catadioptric system image in Fig. 3) designed to directly obtain the soccer field bird's eye view, thus preserving the field geometry in the image. The image green-white-green color transitions over a pre-determined number of circles centered with the robot are collected as the set of transition pixels. The Hough Transform is applied to the set of transition pixels in a given image, using the polar representation of a line (Gonzalez, R., & Woods, R., 1992):

$$\rho = x_i^t . \cos\phi + y_i^t . \sin\phi \qquad (1)$$

where $(x_i^t, y_i^t)$ are the image coordinates of transition pixel $p^t$ and $\rho, \phi$ are the line parameters. The $q$ straight lines

$(\rho_l, \phi_l), \ldots, (\rho_q, \phi_q)$ corresponding to the top $q$ accumulator cells in Hough space are picked and, for all pairs $\{(\rho_j, \phi_j), (\rho_k, \phi_k), j,k=1, \ldots,q, j \neq k\}$ made out of those $q$ straight lines the following distances in Hough space are computed:

$$\Delta\phi = |\phi_j - \phi_k| \quad \Delta\rho = |\rho_j - \rho_k| \qquad (2)$$

Note that a small $\Delta\phi$ denotes almost parallel straight lines, while $\Delta\rho$ is the distance between 2 parallel lines. The $\Delta\phi$ and $\Delta\rho$ values are subsequently classified by *relevance functions* which, based on the knowledge of the field geometry, will filter out lines whose relative orientation and/or distances do not match the actual field relative orientation and/or distances. The remaining lines are correlated, in Hough space, with the geometric field model, so as to obtain the robot posture estimate. An additional step must be taken to disambiguate the robot orientation. In the application to soccer robots, the ambiguity is due to the soccer field symmetry. The goal colors are used to remove such ambiguity and to detect situations where the localization values obtained are not trustable.

Currently, an efficiently coded version of the algorithm is used by each of the ISocRob team robots to obtain its self-localization during a game every second. The algorithm runs in parallel with all the other processes and can compute self-localization in about 13 ms on the average, using Intel IPP library. The knowledge of each robot localization is useful for individual robot navigation, but it is also used by the robot to share information with its teammates regarding team postures and ball location.

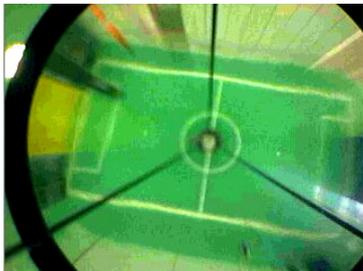

Fig. 3 – Bird's eye-view of the field obtained by the top catadioptric systems of the robots in Fig. 1.

*3.3. Multi-Sensor Guidance with Obstacle Avoidance*
The ability to navigate at relatively high speeds through an environment cluttered with static and dynamic obstacles is a crucial issue for a mobile robot. Most robotic tasks require a robot to move to target postures adequate to carry out its planned activities. In robotic soccer, relevant activities include facing the opponent goal with the ball in between or covering the team goal by positioning itself between the ball and the goal, while avoiding the field walls and the other (stopped and moving) robots. Also relevant is the capability to move towards a given posture while avoiding obstacles and keeping the ball (also known as *dribbling*). A guidance control method for non-holonomic (differential drive) vehicles, using odometry, regularly reset by the vision-based self-localization algorithm described before, was first introduced in (Marques, C., and Lima, P., 2002). The vehicle uses a sonar ring for obstacle avoidance. An alternative guidance method has been introduced in (Damas, B., *et al*, 2002), consisting of a modified potential fields method for robot navigation, especially suited for differential-drive non-holonomic mobile robots. The potential field is modified so as to enhance the relevance of obstacles in the direction of the robot motion. The relative weight assigned to front and side obstacles can be modified by the adjustment of one physically interpretable parameter. The resulting angular speed and linear acceleration of the robot can be expressed as functions of the linear speed, distance and relative orientation to the obstacles. This formulation enables the assignment of angular and linear velocities for the robot in a natural fashion. Moreover, it leads to an elegant formulation of the constraints on angular speed, linear speed and acceleration, that enable a soccer robot to dribble a ball, i.e., to move while avoiding obstacles and pushing the ball without losing it, under severe restrictions to ball holding capabilities. It is shown that, under reasonable physical considerations, the angular speed must be less than a non-linear function of the linear speed and acceleration, which reduces to an affine function of the acceleration/speed ratio when a simplified model of the friction forces on the ball is used and the curvature of the robot trajectory is small.

*3.4. Behavior-Based Architectures*
The basic functional architecture of the SocRob team is organized in three levels of decision and responsibility, similar to those proposed in (Drogoul, A., and Collinot, A., 1998): **individual**, which is responsible for all functionalities that involve only one agent; **relational**, which is responsible for the relationships between the robot and its teammates; and **organizational**, which is responsible for the strategic decisions that involve the team as a whole. The current instantiation of this functional architecture considers that:

there is, at the **organizational level**, a mapping from the environment state, including the team state, to a tactical decision, resulting in an organizational behavior displayed by the team. The tactics consists of the set of *role* assignments to each team member. In robotic soccer, basic *roles* can be `Goalkeeper`, `Defender`, `Attacker` and `Full Player` (both defender and attacker). Only the captain robot will have the organizer enabled. Should the captain "die", the next robot in a pre-specified list will have its organizer level enabled and become the captain.

there are, at the **relational level**, operators which control relations between two or more team members (e.g., to pass a ball, to avoid moving simultaneously towards a ball, to cover a field region while the teammate advances in the field through role exchanges). Any team member has relational operators running. Each operator has a pre-conditions set and, when this set is satisfied, establishes communications with the relational operator(s) of designated teammates, asking them to start a negotiation process which may end up in a coordinated action among this temporary sub-team. As a result, a *relational behavior* is displayed.

there are, at the **individual level**, operators consisting of single *primitive tasks* or of *composite tasks* (primitive tasks linked by logical conditions on events).

The software architecture is the practical implementation of the functional architecture, which could be done in any programming language and using different software technologies. In the SocRob project, the software architecture was defined based on three essential concepts: micro-agents (µA for short), *blackboard* and *plugins*.

Inspired by the idea of Society of Agents, proposed by Minsky (Minsky, M., 1988), each functional module of the SocRob architecture was implemented by a separate process, using the parallel programming technology of threads. In this context a functional module is named µA. In the current implementation of the SocRob architecture there are nine different threads, but only the three most important ones are mentioned here: µA Vision, responsible for processing the data acquired from the cameras, µA Fusion, which fuses information concerning the same object from different sensors, µA Machine, responsible for deciding which behavior should the robot display, and µA Control, responsible for the execution of the corresponding operator.

The concept of threads was chosen to improve module performance and simplify the information passing among the threads. This was accomplished by the blackboard concept (memory space shared by several threads), further sophisticated here by the development of a *distributed blackboard*, in what information availability is concerned. Instead of being centralized in one agent, the information is distributed among all team members and communicated when needed.

As mentioned before, the decision making involved for each agent is twofold: which behavior should be displayed, and how the operator which displays such behavior is executed. This separation between behavior decision and operator execution allows the µA Machine, the one responsible for behavior decision, to work with abstract definitions of behaviors, and choose among them without knowing details about their execution. So, new operators could be easily added and removed without affecting the existing ones, and these can also be easily replaced by others with the simple restriction of maintaining the name. This was accomplished using the concept of *plugin*, in the sense that each new operator is added to the software architecture as a plugin, and therefore the µA Control can be seen as a multiplexer of plugins. Examples of already implemented operators are: `dribble`, `score`, `go`, `standby`, to name but a few. The same idea of plugins was also used for the µA Vision, as each particular functionality related to vision data is defined as a different plugin, and multiplexed by the µA Vision (e.g., a plugin for the front camera, a plugin for the up camera, a plugin for the self-localization algorithm, etc.).

The individual operators have been implemented as state machines, where the states represent primitive tasks, while the arcs between states (if any) are traversed upon the validation of given logical conditions over events (e.g., `see ball`, `distance < x`). The relational operator state machines could also be defined similarly, but events include synchronization signals between the state machines running in the sub-team robots.

However, the way the functional architecture was conceptualized allows the implementation of these operators and the switching among them using different approaches, as for example AI production systems. So, in order to have a more abstract way to deal with behaviour switching, the µA machine has been implemented using a distributed decision-making architecture supported on a logical approach to modeling dynamical systems (Reiter, 2001), based on situation calculus, a first order logic dialect. This architecture includes two main modules: i) a basic logic decision unit, and ii) an advanced logic decision unit. Both run in parallel; the former intends to quickly suggest, using simple logical decision rules, the next behavior to be executed, whereas the latter uses more sophisticated reasoning tools (situation calculus) capable of planning, learning and decision-making, both for individual and cooperative (teamwork) situations. This configures an hybrid architecture where the basic (reactive) unit only controls the robot if the advanced (deliberative) unit takes too long to make a decision, assuming a situation urgency evaluation. A partial implementation of this architecture, the basic logic decision unit, was already performed using Prolog (Arroz, M., *et al*, 2004). Its modeling convenience allowed the quick development of different roles for field players (`Attacker, Defender, Full-Player`), as well as dynamic role change between field players (defenders switch with attackers, depending on who is in a better position to get the ball). The advanced (deliberative) unit, Advanced Logic Based Unit, has been developed using an action programming language called Golog Golog (Levesque, H., *et al*, 1997) and it is based on situational calculus. This unit is responsible to determine plans (sequences of behaviours) that allow the team to achieve something (like scoring on the opposite goal). Situational calculus is an extension to first-order logic, specially suited to handle dynamic worlds. The changes in the world are the results of actions, that have pre-conditions and effects. Our objective is to develop a tool capable of planning and performing task control execution in a distributed environment. To do so we assume that: the agents (robots) can generate, change and execute plans; a plan can be generated, and executed by one or more agents; decisions over the generated plans are based on hypotheses, i.e., assumptions over future states that cannot be guaranteed; and the agents have the capacity to communicate among them, and share information about plans or environment states.

Another recent topic in the project research is the design and implementation of relational behaviors, where teamwork between two or more robots is required to perform a certain task, like a ball pass (Vecht, B., & Lima, P., 2004). These behaviors have a general formulation based on Joint Commitment Theory (Cohen, P. R., & Levesque, H. J., 1991), and use the navigation methods already developed in the project. Currently, the robots are capable of committing to a relational pass behavior where one of the robots is the kicker and the other the receiver. If any of the robots ends the commitment, the other switches to an individual behavior. One cooperation mechanism, implemented in 2000, consists of avoiding that two or more robots from the same team attempt to get the ball. A relational operator was developed to determine which robot should go to the ball and which one(s) should not. In the current implementation, each robot that sees the ball and wants to go for it uses a heuristic function to determine a fitness value. This heuristic penalizes robots that are far from the ball, are between the ball and the opposite goal and need to perform a angular correction to center the ball with its kicking device. Each robot broadcasts its own heuristic value, and the robot with the smallest value is allowed to go for the ball whereas the others execute a `Standby` behavior. Though not tested yet in real robots, formal work on Stochastic Discrete-Event Systems modeling of a multi-robot team has been recently carried out within the project with interesting results (Damas, B., & Lima, P., 2004). The environment space and each player (opponent and teammate) actions are discretized and modeled by a Finite State Automaton (FSA) 2 *vs* 2 players game model. Then, all FSA are composed to obtain the complete model of a team situated in its environment and playing an adversarial game. Controllable (e.g., `shoot_p1`, `stop_p2`) and *Uncontrollable* (e.g., `lost_ball`, `see_ball`) events (i.e., our robots actions) are identified and exponential distributions are assigned to their inter-event times. Dynamic programming is applied to the optimal selection of the *controllable* events, with the goal of minimizing the cost function

$$\min_{\pi} \left[ \int_0^{\infty} C[X(t), u(t)] dt \right] \quad (3)$$

where $\pi$ is a *policy*, $X(t)$ the game state at time $t$, and $u(t)$ is a controllable event, with the cost of *unmarked* states equal to 1, and all the other states have zero cost. If the only marked states are those where a goal is scored for our team, and there are no transitions from marked to unmarked states, this method obtains the minimum (in a stochastic sense) time to goal for our team, constrained by the opponent actions and the uncertainty of our own actions. Some of the chosen actions result in cooperation between the two robots of the team.

**4. Problems To Be Addressed**

Naturally, several interesting problems remain to be tackled and solved within the project research. We will only mention the currently most important ones.

**Behavior Modeling**: A consistent model for individual and relational behaviors is required to provide a systematic methodology for behavior synthesis and analysis. FSA have been used for this purpose up to now. They have the advantage of the availability of several tools for analysis and synthesis in the literature (Cassandras, C. G., & Lafortune, S, 1999), but suffer from limited modeling capabilities. Petri nets (Cassandras, C. G., & Lafortune, S, 1999) extend the modeling capabilities of FSA and provide a more convenient modeling methodology starting from the identification of the system components and events. A wide range of analysis (e.g., concerning boundedness, liveness, stochastic and deterministic time) and synthesis (e.g., concerning admissible marked languages) tools is also available, and the non-decidability of some analysis problems can be overcome with no significant expenses. Furthermore, modularity and system design can be achieved by interconnecting several sub-systems, each modeled as a Petri net. This is particularly convenient to model relational behaviors, where more than one teammate is involved. So, Petri nets are being investigated as an alternative tool for behavior modeling. Behavior switching can also be modeled as discrete-event systems supervision, for which there are results available regarding FSA and Petri nets. Production systems also have modeling characteristics that make them suitable for this purpose. However, further work must be done to study its design and analysis properties.

**Distributed Planning**: The available behaviors among which switching is possible are currently designed "by hand". However, a more appropriate approach would be to develop a planner capable of periodically (or when invoked) analyzing the world state and providing a new set of individual and relational behaviors appropriate for the current conditions. A suitable approach should be the continuous interleaving of plan generation and execution. Task allocation among the team robots and distributed world modeling are relevant issues to be further investigated under this topic.

**Cooperative Learning**: One possible way of designing plans which continuously adapt to new situations and are fine tuned to the actual surrounding environment is to use reinforcement learning (RL) algorithms, especially those which guarantee convergence properties (Sutton, R., and Barto, A., 1998). However, learning is usually slow. An envisaged approach that overcomes this problem is to provide plans with alternative paths among which the RL algorithms can learn to switch over time. Cooperative learning arises when a robot takes its decisions from information learned and provided to it by its teammates.

**Control as a Game**: Modern views of control state the control problem as a game against an adversary (i.e., the disturbances). In the particular case of soccer, there is an actual opponent whose modeled behavior, once estimated (e.g., using Hidden Markov Models), can be used as information for game-playing algorithms, as part of the planning process.

## 5. Conclusion

This paper described the SocRob project (on the development of methodologies for analysis, design and implementation of multi-robot cooperative systems), its objectives, past, current and intended future work. One interesting feature of the project is that it enables different approaches to the solution of the problem at hand. This naturally motivates competing research approaches, as well as research on analysis methods to compare the different results. Furthermore, the project fosters education in AI and Robotics related topics, because so many issues must be solved to handle the overall problem. Students from different levels (undergraduate, graduate, post-doctorate) can get involved at different difficulty levels and accomplish project sub-goals. They also learn how to accomplish teamwork under hard time deadlines. The SocRob project has involved so far 10 undergraduate and 4 graduate (MSc and PhD) students, besides 2 doctorates who have been supervising the project. All these students have participated regularly in RoboCup - The World Cup of Soccer Robots, since 1998. We believe that RoboCup is a very attractive long-term scientific challenge that brings together people from several different scientific fields in an exciting fusion of research, education and science promotion which are actually the driving forces of our project too.

Some of the methodologies developed within the project, namely its software and functional architectures, have been applied meanwhile to other projects, such as an European Space Agency project on Formation Guidance and Navigation of Distributed Spacecraft, and a Cooperative Navigation for Rescue Robots project currently underway at ISR/IST.

The project team is now developing new robots, in the framework of a national research project, in partnership with two Portuguese small companies. These new robots are omnidirectional, with a new modular construction, so that it will be easily modified, e.g., the up camera module can switch between a catadioptric system and a stereo image system. The new robots will also incorporate a controlled kicker mechanism, so that one can choose the kicking force, using an electromechanical solution with a DC motor pulling a spring and an infrared sensor to measure the pulled distance, both coupled to the kicking device. In order to make new and more complex behaviors and for ball handling, there is a ball reception mechanism, that will allow the implementation of ball passes behaviors. Two new sensors will be used: a rate-gyro for angular velocity measurements, and an optical mouse to track the robot position in the field. Both will provide data to be fused with odometry and vision-based self-localization, so as to improve navigation.